\documentclass[preprint,12pt]{elsarticle}

\usepackage{amssymb}
\usepackage{amsmath}
\usepackage{algorithm}
\usepackage{algorithmic}
\usepackage{booktabs}
\usepackage{graphicx}
\usepackage{stfloats}
\usepackage{url}
\usepackage{tikz}
\usetikzlibrary{arrows.meta, positioning, calc, backgrounds, fit}

\journal{Automation in Construction}

\begin{document}

\begin{frontmatter}

\title{BIM Information Extraction Through LLM-based Adaptive Exploration}

\author[tum]{S.~Hellin\corref{cor1}}
\ead{sylvain.hellin@tum.de}
\author[tum]{S.~Jang}
\author[tum]{S.~Fuchs}
\author[tum]{S.~Nousias}
\author[tum]{A.~Borrmann}

\cortext[cor1]{Corresponding author}

\affiliation[tum]{organization={Chair of Computing in Civil and Building Engineering, Georg Nemetschek Institute, Technical University of Munich},
    city={Munich},
    country={Germany}}

\begin{abstract}
BIM models provide structured representations of building geometry, semantics, and topology, yet extracting specific information from them remains remarkably difficult. Current approaches translate natural language into structured queries by assuming a fixed data organization (static approach), which BIM heterogeneity eventually invalidates.

We address this with a new paradigm, adaptive exploration, where an LLM-based agent iteratively executes code to extract information from a BIM model, discovering its structure at runtime instead of assuming it.

We evaluate this approach on ifc-bench~v2, an open-source BIM question-answering benchmark introduced alongside this work, comprising 1,027 tasks across 37 IFC models from 21 projects.

A factorial ablation across two LLM capability levels and four augmentation strategies shows that adaptive exploration significantly outperforms static query generation across all configurations, regardless of the augmentation strategy. These results indicate that BIM heterogeneity is best addressed at the paradigm level, not by further optimizing static approaches.
\end{abstract}




\end{frontmatter}


\section{Introduction}
\label{sec:introduction}

As the construction industry increasingly relies on Building Information Modeling (BIM), the ability to programmatically extract specific information from BIM models becomes a critical bottleneck~\cite{borrmannBuildingInformationModeling2018, wangBIMHandbookGuide2012}. These models encode detailed representations of a building's geometry, semantics, and topology, yet most BIM tools are designed for data creation and visualization, not for intuitive data consumption~\cite{weiTexttostructureInterpretationUser2025, olofssonhallenInteractionsHumanTechnology2023}. This creates a persistent barrier between the information stored in BIM models and the downstream applications that depend on it, from facility operations and construction management to compliance checking.

Bridging this barrier requires overcoming two interrelated challenges. The first is a \emph{query expertise gap}: the domain experts who need information from BIM models -- architects reviewing design compliance, construction managers tracking quantities, facility operators locating equipment -- are rarely trained in the query languages or API calls required to retrieve it. Most end users lack programming skills entirely, and even BIM-proficient professionals struggle to extract non-trivial information such as cross-element aggregations or derived quantities~\cite{dongAIBIMCoordinator2025}. Existing graphical interfaces offer limited query expressiveness: they support predefined filters and property lookups but cannot handle arbitrary aggregation or computation tasks. This information access barrier is a recognized obstacle to broader BIM adoption, because the value of richly modeled data remains unrealized when stakeholders cannot retrieve the information they need~\cite{wangBIMHandbookGuide2012}. Crucially, the person retrieving information is often not the original model author, and thus lacks knowledge of the specific modeling decisions that shaped the data.

The second challenge is the structural heterogeneity of BIM data (hereafter, \emph{BIM heterogeneity}). Even within a single vendor ecosystem, models differ in how information is represented. A property like gross floor area may be stored as an explicit attribute or derived from geometry; a door width may appear as \texttt{Width}, \texttt{Rough Width}, \texttt{NominalWidth}, or \texttt{Breite~(B)} depending on the authoring tool, modeling convention, and language (Section~\ref{subsubsec:handling-heterogeneity}). Such variation persists even though the IFC schema itself provides standardized attributes (e.g., \texttt{OverallWidth}) and dedicated property sets (e.g., \texttt{IfcDoorLiningProperties}) since its earliest versions, and homogenization efforts such as the buildingSMART Data Dictionary (bSDD) or the German BIM portal target exactly this problem. It is not an edge case; it is the norm in practice.

The Industry Foundation Classes (IFC) standard, the predominant open data format for exchanging BIM data across vendors and tools, amplifies this heterogeneity by design: adopted by virtually every major BIM software vendor, IFC must accommodate diverse modeling paradigms and therefore permits wide variation in how information is structured~\cite{borrmannIndustryFoundationClasses2018, handykosasihBIMQualityControl2024}. The underlying challenge, however, is general. For any system intended to handle diverse, real-world BIM models without enforcing specific modeling conventions, approaches that assume a known or consistent data structure will systematically break down.

\begin{figure}[htbp]
\centering
\includegraphics[width=\columnwidth]{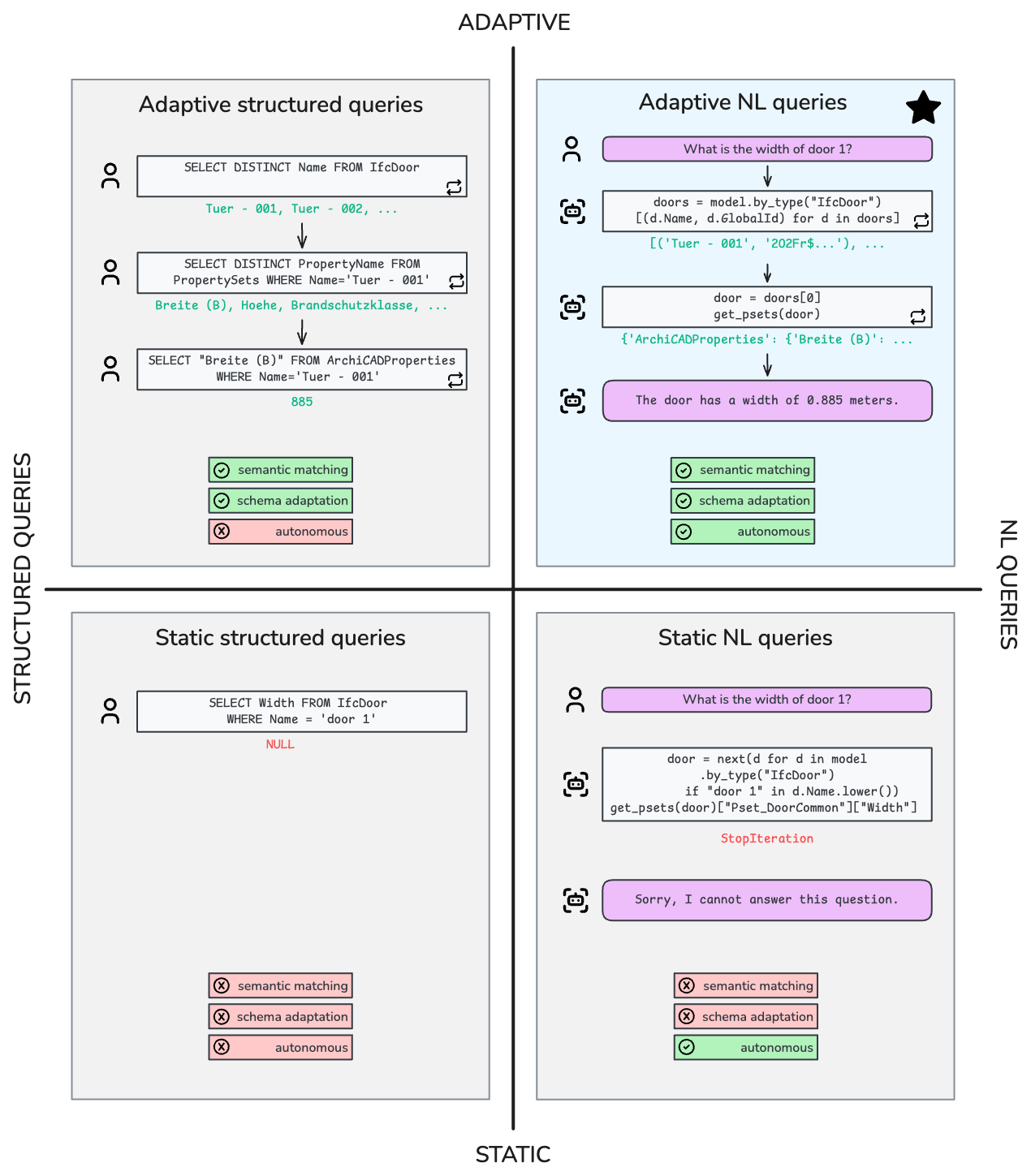}
\caption{Four approaches to BIM information extraction, distinguished by how users formulate their queries (structured vs natural language) and execution strategy (static vs dynamic). The same question about door width yields different outcomes depending on whether the approach can handle naming variation (semantic matching) and structural variation (runtime data discovery). Only adaptive exploration (top-right) addresses both dimensions.}
\label{fig:query-paradigms}
\end{figure}

These two challenges -- query expertise and BIM heterogeneity -- interact: a system must both accept natural language input (bridging the expertise gap) and adapt to each model's data layout at runtime (handling heterogeneity). Existing approaches address at most one dimension, as illustrated in Figure~\ref{fig:query-paradigms}.

Conventional approaches translate a natural language query into a single structured query in languages such as SQL or SPARQL~\cite{guoApproachAutomaticSPARQL2020, liuIntegratedMethodBIM2025}. These methods require converting BIM models into alternative representations that typically capture only a subset of the full IFC schema. The limitation is the fixed, design-time assumptions about data structure: these systems support only pre-defined query types with exact name matches and cannot adapt to heterogeneous data at runtime.

While these limitations motivated exploration of AI-driven alternatives, LLM-based approaches have so far brought only partial progress. They introduced some level of semantic understanding to BIM information extraction, yet all retain static architectures: code generation without iterative execution, single-pass pipelines on pre-extracted data, or pre-determined workflows. None iteratively adapts its exploration strategy based on what it discovers in the actual data at runtime. Section~\ref{sec:related-work} reviews these approaches in detail.

In our prior work~\cite{hellinNaturalLanguageInformation2025}, we took a first step toward runtime adaptation by applying an LLM-based system to BIM information extraction with a fixed set of 29 manually coded tools (pre-built Python functions such as \texttt{get\_elements\_by\_type}). The system could reason iteratively, yet its actions were bounded by the hand-crafted toolset that proved impractical to extend; tool limitations accounted for 45\% of errors (Section~\ref{subsubsec:dynamic-exploration-rw}).

This work removes this constraint through \textbf{adaptive exploration}. Rather than relying on pre-built tools, the agent writes and executes arbitrary Python code against data models, iterating based on execution results and adapting its strategy at runtime. We use the term \emph{agent} to denote an LLM that operates in a loop, taking actions and receiving feedback from its environment until it achieves its goal~\cite{wangSurveyLargeLanguage2024a, zhu2025establishing}. We evaluate this approach across model capability levels and \emph{augmentation strategies}, that is, supplementary resources such as API documentation or domain-specific tools provided alongside the base agent to potentially improve extraction accuracy, on diverse unseen BIM models.

Among existing LLM-based approaches, static code generation with single-pass execution~\cite{austernComparingDifferentBuilding2025} represents the most capable paradigm (Section~\ref{subsubsec:llm-approaches}), making it the strongest available baseline for isolating the effect of adaptive exploration. Beyond the exploration paradigm itself, augmentation strategies such as documentation retrieval and domain-specific tools can supplement the base agent, raising the question of how these interact with LLM capability.

These considerations motivate two research questions:

\begin{itemize}
\item[\textbf{RQ1:}] How effective is adaptive exploration compared to static code generation for BIM information extraction?
\item[\textbf{RQ2:}] How do augmentation strategy and language model capability interact in determining the effectiveness of adaptive exploration?
\end{itemize}

In answering these questions, this work makes the following contributions.

\begin{enumerate}
\item An adaptive exploration paradigm for BIM information extraction, validated across two model capability levels. Adaptive exploration consistently and significantly outperforms static code generation (Section~\ref{subsec:rq1}), with the paradigm choice dominating all other factors in the ablation.

\item ifc-bench v2, the first large-scale, open, permissively licensed BIM question-answering benchmark (1,027 tasks across 37 IFC models from 21 projects), enabling broad, reproducible evaluation of LLM-based BIM information extraction. Unlike prior datasets that contain only deterministically verifiable answers, ifc-bench v2 includes open-ended questions requiring judgment and estimation, reflecting the broader landscape of real-world information needs.

\item Systematic evaluation demonstrating that augmentation effectiveness is strongly model-capability-dependent, based on a factorial ablation across four augmentation strategies and two model capability levels, challenging prior work~\cite{zhouDocPromptingGeneratingCode2022, caiLargeLanguageModels2024}: the stronger model is invariant to augmentation ($\pm$0.6pp across all strategies, all $p > 0.8$), while documentation significantly improves the weaker model (+4.9pp, $p < 0.05$).

\end{enumerate}

The remainder of this paper is organized as follows. Section~\ref{sec:related-work} reviews related work across BIM extraction paradigms, augmentation strategies, and evaluation methodology. Section~\ref{sec:methodology} presents the agent architecture, documentation retrieval system, and automated tool generation pipeline. Section~\ref{sec:experimental-setup} describes ifc-bench v2, the evaluation framework, and experimental design. Section~\ref{sec:results} reports results organized by research question, Section~\ref{sec:discussion} discusses mechanisms, implications, and limitations, and Section~\ref{sec:conclusion} provides the conclusions.

\section{Related Work}
\label{sec:related-work}

Research on BIM information extraction has evolved through two main trajectories: query translation approaches and LLM-based approaches. This section reviews existing work along both trajectories, including the state of benchmarks and evaluation methodology, and then discusses augmentation strategies for code-based agents (Section~\ref{subsec:augmentation-strategies}).

\subsection{Existing Approaches to BIM Information Extraction}
\label{subsec:bim-ie}

\subsubsection{Query Translation Approaches}
\label{subsubsec:query-translation}

The earliest paradigm translates natural language into structured query languages. These approaches require converting IFC models into alternative representations: relational databases for SQL~\cite{liuIntegratedMethodBIM2025, shinBIMASRFrameworkVoiceBased2021}, RDF/OWL graphs for SPARQL~\cite{guoApproachAutomaticSPARQL2020, pauwelsEXPRESSOWLConstruction2016}, or noSQL databases such as MongoDB~\cite{linNaturalLanguageBasedApproachIntelligent2016}. Each conversion requires significant manual preprocessing effort, and the resulting representation constrains the system to query types and element categories explicitly accounted for during system development. Other approaches extract subsets of BIM data into simplified structures: CSV files for voice assistants~\cite{elghaishArtificialIntelligencebasedVoice2022} or hierarchical trees for multi-scale retrieval~\cite{wangMultiscaleInformationRetrieval2021}.

Classical NLP and ML techniques extend query translation further. Wang et al.~\cite{wangNLPBasedQueryAnsweringSystem2022} report 81.9\% accuracy across 11 query types, though this metric reflects performance on a narrow, pre-defined query set rather than general extraction capability. Ontology-aided approaches~\cite{yinOntologyaidedNaturalLanguagebased2023a, yinTwostageTexttoBIMQLSemantic2023} use domain ontologies and semantic parsing to reduce sensitivity to naming variation, but remain constrained to queries within the pre-defined ontology coverage.

All query translation approaches share fundamental limitations: they require manual preprocessing, support only pre-defined query types with exact name matches, and cannot generalize to query or element types beyond those explicitly accounted for during system development. This narrow coverage is a fundamental constraint: each approach targets a fixed inventory of pre-defined query patterns and fails on any question outside this scope.~\cite{handykosasihBIMQualityControl2024}. Despite extensive efforts to enable natural language querying -- including the use of LLMs -- handling heterogeneous BIM data remains a major challenge, as most approaches continue to operate within a static translation paradigm.

\subsubsection{LLM-Based Approaches}
\label{subsubsec:llm-approaches}

LLMs brought some level of semantic understanding to BIM extraction, yet existing systems retain static architectures that prevent runtime adaptation. We organize these approaches by the key limitation each shares.

\paragraph{Code generation without execution feedback}
Guo et al.~\cite{guoAdvancingBIMInformation2025} and Koh et al.~\cite{kohCosteffectiveMinimalinterventionBIM2026} generate Revit C\# code through multi-agent pipelines; the pipelines produce code that must be executed separately to obtain results, not direct answers to user questions. Guo et al.\ report 78.75\% alignment accuracy with 8 agents, though generated code ``could not be executed directly due to minor mistakes''~\cite{guoAdvancingBIMInformation2025}. Koh et al.\ extend this pipeline with a 6-checker suffix module, achieving 37.50\% semantic accuracy (SemAcc), the proportion of queries where generated code both executes successfully and returns the intended output~\cite{kohCosteffectiveMinimalinterventionBIM2026}. The authors acknowledge that ``this approach inherently limited direct interaction between the LLM and the execution environment''~\cite{kohCosteffectiveMinimalinterventionBIM2026}. Both approaches require standardized Revit models with exact name matches.

\paragraph{Static pipelines on pre-extracted data}
Several approaches extract BIM data into intermediate formats before querying. Gao et al.~\cite{gaoDomainSpecificFineTuningPromptBased2025} combine intent classification with table question-answering on pre-extracted CSV files, supporting three aggregation functions. Zheng and Fischer~\cite{zhengDynamicPromptbasedVirtual2023} first parse a subset of the BIM model into MongoDB, then chain GPT prompts, achieving partial value-level matching by injecting all unique parameter values into prompts, but the pipeline is fixed and limited to direct attribute queries. Li and Wang~\cite{liBuildingGPTQueryBuilding2026} generate SPARQL queries for Brick Schema via vector-graph retrieval-augmented generation, surpassing 90\% accuracy across four question types; Li et al.~\cite{liEnhancingLLMbasedBuilding2026} extend this with fine-tuning, reaching 97.11\% with a fine-tuned LLaMA~3.1-70B. This high accuracy reflects Brick's standardized ontology, a controlled setting where the schema is known and consistent, contrasting with heterogeneous IFC models.

\paragraph{Code generation with single-pass execution}
Unlike the approaches above, which produce code without executing it (and therefore do not attempt to answer the question directly), the following work generates code, executes it in a single pass, and uses the output to produce an answer. Austern et al.~\cite{austernComparingDifferentBuilding2025} compare BIM representations (JSON, graph, IFC, Revit API) for LLM-based extraction. Their approach generates code, executes it, and feeds errors back for syntax correction. This is the closest existing work to our static baseline and serves as the reference paradigm for our comparison. Their key finding, that ``reliance on data-schemas which might not be implemented in the BIM model is a serious drawback for LLM based methods''~\cite{austernComparingDifferentBuilding2025}, directly motivates adaptive exploration.

LLMs have also been applied to adjacent BIM tasks such as coordination~\cite{dongAIBIMCoordinator2025} and authoring~\cite{leeGeneralizedLLMAugmentedBIM2024}, but these address different problems than information extraction. Despite varying sophistication, all static LLM-based extraction approaches share a fundamental constraint: none can adapt its exploration strategy at runtime based on what it discovers in the data.

\paragraph{Agentic approaches}
\label{subsubsec:dynamic-exploration-rw}
In our prior work~\cite{hellinNaturalLanguageInformation2025}, we introduced a first agentic approach to BIM information retrieval. Using a ReAct agent~\cite{yaoReActSynergizingReasoning2022} with 29 manually coded tools, the system achieved 80\% accuracy on 99 queries while operating directly on IFC files without pre-processing. The approach brought multi-step reasoning and iterative tool use to BIM extraction: variable-length reasoning chains, diverse query types, and semantic matching through repeated interaction with the data.

However, the agent's capabilities were bounded by its fixed toolset. Error analysis revealed that 45\% of failures originated from tools: 25\% from incorrect implementations and 20\% from missing tools. Tool creation proved labor-intensive and error-prone, making it impractical to develop comprehensive toolsets for all potential query types.

Concurrently with this work, Gao et al.~\cite{gaoMultiagentFrameworkSchemaguided2026} introduced a multi-agent framework with schema-guided tool sequencing for IFC interaction. Their system dynamically composes sequences of pre-built atomic tools at runtime, switching between per-entity agent dispatch and a sample-and-generalize execution mode for scalability. While this represents an advance over single-pass approaches, the agent's capabilities remain bounded by its hand-crafted toolset: it selects from pre-defined functions rather than generating arbitrary code, and critically, cannot adapt its exploration strategy based on runtime feedback from the execution environment. Their scalability mode further assumes schema homogeneity across entities of the same type, a simplification that does not hold across heterogeneous real-world models.

Both agentic approaches share the same fundamental constraint: exploration is limited to pre-defined tools. The CodeAct architecture (Section~\ref{subsubsec:codeact}) provides the foundation for removing this constraint: by replacing tool-based actions with executable code, the agent is no longer bounded by a pre-defined action space.

\subsubsection{Benchmarks and Evaluation}
\label{subsubsec:benchmarks-eval}

Existing BIM extraction evaluations rely on small, non-standardized test sets that prevent cross-study comparison. Wang et al.~\cite{wangNLPBasedQueryAnsweringSystem2022} evaluate on 11 queries, Guo et al.~\cite{guoAdvancingBIMInformation2025} and Koh et al.~\cite{kohCosteffectiveMinimalinterventionBIM2026} both used the same non-public dataset of 80 queries, and our prior work~\cite{hellinNaturalLanguageInformation2025} released the first open-source BIM-QA dataset with 99 queries (ifc-bench v1). Each study defines custom metrics on private or non-standardized datasets, making cross-comparison impossible. No large-scale, open benchmark exists for BIM information extraction, a gap analogous to the role SWE-bench fills for software engineering tasks.

Beyond benchmark scale, evaluation methodologies in BIM information extraction present two persistent problems: binary accuracy masks quality variation by obscuring systematic failures on specific question types, and custom metrics on private datasets prevent cross-study comparison. In prior work~\cite{hellinsylvainSystematicEvaluationFramework2026, hellinsylvainENABLINGCROSSSTUDYCOMPARISON2026}, we developed a unified evaluation framework for BIM extraction: a multi-criteria taxonomy that captures quality variation beyond binary accuracy, coupled with an LLM-as-judge protocol validated across two LLM families ($\alpha = 0.70$--$1.00$). The LLM judge receives the original question, the ground-truth answer, the system's answer, and the full execution trace, then evaluates each of the five quality criteria independently according to category-specific rubrics. This framework underpins the evaluation in Section~\ref{sec:experimental-setup}.

\subsubsection{Synthesis}
\label{subsubsec:synthesis}

BIM information extraction has advanced from query translation, limited to pre-defined query types and element categories, to LLM-based approaches with semantic understanding, and agentic approaches with iterative reasoning. Recent agentic systems~\cite{hellinNaturalLanguageInformation2025, gaoMultiagentFrameworkSchemaguided2026} dynamically sequence pre-built tools at runtime, yet none is truly adaptive: the agent's action space remains bounded by its hand-crafted toolset.

Three dimensions of heterogeneity structure the remaining gaps, each requiring a distinct capability:

\begin{enumerate}
\item \emph{Schema-level adaptation}: how is information organized in a given model? A property like gross floor area may reside as an explicit attribute or must be derived from geometry. Static approaches assume a fixed schema and break when data is organized differently. This maps onto the vertical axis of the query paradigm framework (Figure~\ref{fig:query-paradigms}): dynamic approaches (top row) discover the actual data layout at runtime.

\item \emph{Key-level semantic matching}: once the relevant attribute collection is located, what is the property called? A width property may appear as \texttt{Width}, \texttt{NominalWidth}, or \texttt{Breite~(B)} depending on the authoring tool and language. Ontology-aided methods~\cite{yinOntologyaidedNaturalLanguagebased2023a} and key-level matching~\cite{guoAdvancingBIMInformation2025} address this dimension to varying degrees.

\item \emph{Instance-level entity resolution}: matching user references to specific entity instances in the model. While entity type filtering (e.g., retrieving all \texttt{IfcDoor} instances) works regardless of language or modeling convention, identifying a specific entity instance by name requires matching the user's reference against arbitrary string labels stored in the model. A user asking about ``door~1'' must match an entity labeled ``Tuer~-~001'' in a German ArchiCAD model; a query about levels in a Norwegian model must map the user's notion of ``ground floor'' to a level named ``u.etg'' (hitos project). No existing approach handles this resolution.\end{enumerate}

The benchmark corpus spans 37 IFC models from 21 projects, exported from diverse BIM authoring tools (Revit 2011--2025, ArchiCAD 11--25, Synchro, DDS-CAD) with varying export settings, and includes real-world projects.

To illustrate how these compound: asking ``What is the width of door~1?'' on a German ArchiCAD model requires resolving the entity name (value-level), locating the property in \texttt{ArchiCADProperties} instead of the expected \texttt{Pset\_DoorCommon} (schema-level), and interpreting the property name \texttt{Breite~(B)} as ``width'' (key-level). No existing approach handles all three dimensions. Adaptive exploration addresses them simultaneously through iterative code execution against the underlying BIM model. Table~\ref{tab:approach_comparison} summarizes these distinctions across paradigms.

\begin{table}[htbp]
\centering
\small
\caption{Comparison of BIM information extraction approaches across capability dimensions. Runtime adaptation is distinguished from single-pass code execution: it requires iterative strategy adaptation based on discovered data, not merely executing pre-generated code.}
\label{tab:approach_comparison}
\resizebox{\textwidth}{!}{%
\begin{tabular}{l c c l c}
\hline
\textbf{Reference} & \textbf{Prepr.} & \textbf{Runtime} & \textbf{Semantic} & \textbf{Heter.} \\
 & \textbf{req.} & \textbf{adapt.} & \textbf{matching} & \textbf{support} \\
\hline
\multicolumn{5}{l}{\textit{Query translation}} \\
Guo et al.\ (2020)       & Yes & --      & None      & No  \\
Liu et al.\ (2025)       & Yes & --      & None      & No  \\
Shin et al.\ (2021)      & Yes & --      & None      & No  \\
\cline{1-5}
\multicolumn{5}{l}{\textit{LLM-based static}} \\
Guo et al.\ (2025)       & Yes & No      & Key-level & No  \\
Koh et al.\ (2026)       & Yes & No      & Key-level & No  \\
Austern et al.\ (2025)   & Varies & No      & None      & Partial \\
\cline{1-5}
\multicolumn{5}{l}{\textit{LLM-based agentic}} \\
Hellin et al.\ (2025)    & No  & Partial$^{a}$ & Key-and-value & Yes \\
Gao et al.\ (2026)       & No  & No$^{b}$      & Key-level & Partial$^{c}$ \\
\cline{1-5}
\multicolumn{5}{l}{\textit{Adaptive exploration}} \\
\textbf{This work}        & \textbf{No}  & \textbf{Yes} & \textbf{Key-and-value} & \textbf{Yes} \\
\hline
\end{tabular}}%
\begin{flushleft}
\footnotesize
$^{a}$\,Iterative tool use with strategy adaptation, but constrained to 29 pre-defined tools; cannot generate code to explore arbitrary schema paths. \\
$^{b}$\,Dynamic tool sequencing via ReAct, but constrained to pre-built atomic tools; no arbitrary code generation or structural adaptation based on runtime feedback. \\
$^{c}$\,Mode~II assumes schema homogeneity across entities of the same type. \\
\textit{Columns:} Prepr.\ req.\ = format conversion before querying. Runtime adapt.\ = iterative exploration with strategy adaptation based on intermediate results, including structural adaptation to varying data organizations and schema layouts. Semantic matching: None = exact string match; Key-level = property name mapping; Key-and-value = matching on both property names and data values. Heter.\ support = works on heterogeneous, unstandardized models.
\end{flushleft}
\end{table}

\subsection{Augmentation Strategies for Code-Based Agents}
\label{subsec:augmentation-strategies}

Since adaptive exploration relies on an agent generating and executing code, we first review the architectural foundation (CodeAct) and then survey two augmentation strategies that improve code-based agents: documentation retrieval and domain-specific tool generation.

\subsubsection{From ReAct to CodeAct}
\label{subsubsec:codeact}

ReAct~\cite{yaoReActSynergizingReasoning2022} established the paradigm of interleaving reasoning with tool-based actions, where tools define the agent's entire action space. CodeAct~\cite{wangExecutableCodeActions2024} shifts this architecture by using code as the action space, achieving approximately 30\% fewer interaction steps and 6--20\% higher success rates. This shift has a critical implication for augmentation: with CodeAct, tools become performance boosters rather than requirements, a property impossible under ReAct where tools define the action space.

\subsubsection{Documentation-Augmented Code Generation}
\label{subsubsec:doc-augmentation}

Zhou et al.~\cite{zhouDocPromptingGeneratingCode2022} demonstrate that retrieving relevant documentation during code generation significantly improves performance, with especially large gains for unseen functions (18.30 vs.\ 9.03 recall). Chen et al.~\cite{chenWhenLLMsMeet2025} study retrieval-augmented generation with API documentation for less common Python libraries across 1,017 APIs. Code examples emerge as the critical component; removing them drops accuracy from 0.66--0.82 to 0.22--0.39. Together, these works establish strong evidence that documentation retrieval should improve code-generating agents.

\subsubsection{Domain-Specific Tool Generation}
\label{subsubsec:tool-generation}

LLM-based tool generation has demonstrated consistent benefits across domains. Cai et al.~\cite{caiLargeLanguageModels2024} show that LLMs generate reusable tools from a few demonstrations, enabling inference to shift to a model with 15$\times$ lower per-call API cost while preserving performance. Stengel-Eskin et al.~\cite{stengel-eskinReGALRefactoringPrograms2024} refactor programs into reusable libraries, achieving approximately 26\% improvement on date-understanding tasks. Huang et al.~\cite{huangAgentCoderMultiAgentbasedCode2024} report improvements in code generation accuracy when using multi-agent architectures with independent test generation compared to single-agent baselines. Other frameworks (HuggingGPT~\cite{shenHuggingGPTSolvingAI2023}, Gorilla~\cite{patilGorillaLargeLanguage2023}, Toolformer~\cite{schickToolformerLanguageModels2023}) further establish the feasibility of automated tool creation. TroVE~\cite{wangTroVEInducingVerifiable2024} additionally reports 31\% faster human verification of generated toolboxes, though compute-matched re-evaluation~\cite{sesterhennComputeMatchedReEvaluationTroVE2025} suggests its accuracy gains may reflect higher compute allocation rather than the toolbox mechanism itself. However, none of these approaches has been applied to structured data extraction domains such as BIM.

\section{Proposed Method}
\label{sec:methodology}

We implement this adaptive exploration paradigm as an LLM-based agent: given a natural language question and a BIM model, it iteratively generates and executes code, observes results, and refines its strategy until it can answer (Section~\ref{subsec:exploration-agent}). Two optional augmentation strategies extend this paradigm: documentation retrieval injects relevant API documentation into the agent's context at query time (Section~\ref{subsec:doc-retrieval}), while domain-specific tools provide pre-built abstractions for common extraction patterns, either manually developed or automatically generated through a multi-agent system pipeline (Section~\ref{subsubsec:auto-tool-gen}).

\subsection{Adaptive Exploration Agent}
\label{subsec:exploration-agent}

\subsubsection{Agent Architecture and Execution Loop}
\label{subsubsec:agent-architecture}

The agent follows a CodeAct architecture~\cite{wangExecutableCodeActions2024}, iteratively writing and executing code against live BIM models. At each iteration $t$, the agent receives the question $q$, the model path $m$, the accumulated execution history $H_{t-1}$ (pairs of generated code and execution outputs), and optional tool descriptions $\mathcal{T}$. It produces one of two response types: a \textsc{CodeAction} containing reasoning and Python code, or a \textsc{FinalAnswer} containing reasoning and the extracted answer.

\begin{figure}[htbp]
\centering
\includegraphics[width=\columnwidth]{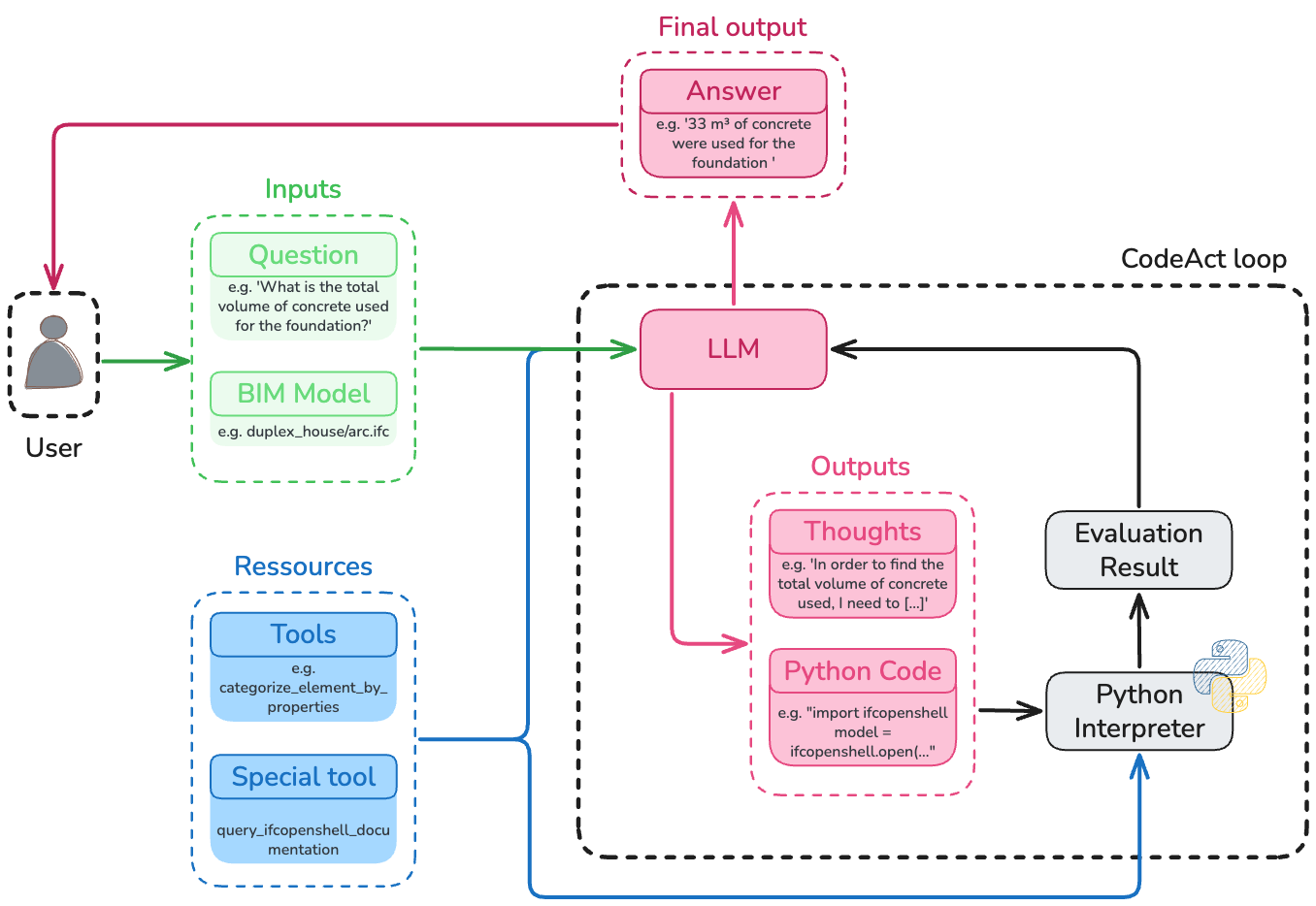}
\caption{Adaptive exploration execution flow. The agent iteratively writes and executes code against the BIM model, receiving execution feedback at each step until it produces a final answer or exhausts its iteration budget}
\label{fig:cobbie-inference}
\end{figure}

Code execution occurs in a sandboxed runtime environment with programmatic access to the BIM model and any available tools. Execution outputs (return values, print output, error messages) are captured and appended to $H$, forming the feedback signal for subsequent iterations. A maximum iteration limit $N$ prevents unbounded exploration; if the agent cannot answer within $N$ iterations, it automatically abstains. Algorithm~\ref{alg:inference} formalizes this loop and Figure~\ref{fig:cobbie-inference} illustrates the execution flow.

The system prompt instructs the agent to answer the user's question about the provided BIM model using iterative code exploration via the available runtime and any provided tools. It directs the agent to prefer higher-level helper functions over direct API calls when available, and to thoroughly explore all relevant element types and properties before answering. Experiment-specific prompt details (answer sourcing hierarchy, quality criteria) are described in Section~\ref{subsec:implementation}.

\begin{algorithm}[htbp]
\caption{Adaptive Exploration}
\label{alg:inference}
\begin{algorithmic}[1]
\REQUIRE Question $q$, BIM model path $m$, tool descriptions $\mathcal{T}$, max iterations $N$
\ENSURE Extracted answer or abstention message
\STATE $H \gets \emptyset$ \COMMENT{Initialize execution history}
\STATE $t \gets 0$
\WHILE{$t < N$}
	\STATE $response \gets \textsc{Agent}(q, m, H, \mathcal{T})$
	\IF{$response = \textsc{FinalAnswer}$}
		\RETURN $response.answer$
	\ELSE
	\STATE $code \gets response.code$
	\STATE $output \gets \textsc{Execute}(code, m)$ \COMMENT{Sandboxed code execution}
	\STATE $H \gets H \cup \{(code, output)\}$
	\STATE $t \gets t+1$
	\ENDIF
\ENDWHILE
\RETURN ``Information not found in BIM model''
\end{algorithmic}
\end{algorithm}

\subsubsection{Handling Heterogeneity Through Exploration}
\label{subsubsec:handling-heterogeneity}

Adaptive exploration addresses BIM heterogeneity through two mechanisms: runtime discovery of data structures and error recovery via execution feedback. Rather than assuming known schemas, the agent discovers how data is structured in each model through iterative code execution.

Consider a query about door widths. Across ifc-bench models, this property appears as \texttt{Width} in Revit models, \texttt{Rough Width} in others, \texttt{NominalWidth} in clinical facility models, and \texttt{Breite (B)} in German ArchiCAD models.\footnote{Observed via manual inspection of IFC property sets: \texttt{Width} in models 4351, dental\_clinic, and hitos (Revit); \texttt{Rough Width} in 4351 and city\_house\_munich; \texttt{NominalWidth} in dental\_clinic; \texttt{Breite (B)} in ac20 and fzk\_house (ArchiCAD).} A static approach that assumes any single naming convention will fail on the rest. The agent discovers the actual property location: it queries the model, observes what property sets exist, inspects their contents, and adapts accordingly.

Beyond schema-level adaptation, iterative exploration enables two further capabilities: value-level semantic matching through the LLM's multilingual understanding, and on-demand geometric computation for numerical properties not stored explicitly in the model. According to the execution traces (the log of code actions and outputs from each iterative agent run), the agent matches the English query term ``outer walls'' to the German value ``Aussenwande'' in an ArchiCAD model's property, and recognizes Norwegian level names (``kjeller'' as basement, ``u.etg'' as ground floor) in Scandinavian models, mapping them to English architectural concepts in the answer. No explicit translation modules are required; the LLM bridges language gaps during iterative exploration. These capabilities align with findings from Self-Refine~\cite{madaanSelfRefineIterativeRefinement2023} and Reflexion~\cite{shinnReflexionLanguageAgents2023}: providing execution feedback to LLMs consistently improves performance, particularly when feedback comes from deterministic tools.

\subsubsection{Role of Augmentation (Tools and Documentation)}
\label{subsubsec:role-augmentation}

The agent can leverage tools and documentation, but retains full code execution access regardless of what augmentation is provided. Tools serve as domain-specific abstractions that reduce code complexity and embed domain knowledge, while documentation provides API reference that reduces the risk of the LLM fabricating calls to non-existent library functions. Critically, the agent is never restricted to tools; it can write arbitrary code when tools are insufficient or unavailable, and embed tools in its written code logic.

This flexibility distinguishes our approach from prior ReAct-based approaches~\cite{hellinNaturalLanguageInformation2025} where tools defined the entire action space. In the CodeAct architecture, tools are performance boosters rather than requirements: the agent retains full code execution access regardless of what augmentation is provided, functioning without any tools or documentation.

\subsection{Documentation Retrieval System}
\label{subsec:doc-retrieval}

Because the agent generates code against external libraries, it benefits from access to API documentation at query time. Rather than injecting full documentation into the prompt, the agent retrieves only the most relevant fragments through a hybrid retrieval pipeline. The pipeline is library-agnostic: given any API documentation corpus, it indexes and serves relevant chunks on demand.

\subsubsection{Documentation Corpus and Processing}
\label{subsubsec:doc-corpus}

The documentation processing pipeline is library-agnostic: given any API documentation corpus (source code, tutorials, reference pages), it produces an indexed collection of retrievable chunks. Chunking is semantic rather than fixed-size: source files are parsed via Abstract Syntax Tree (AST) extraction, where each function, class, or method yields a self-contained chunk containing its name, signature, and docstring. Tutorial and documentation files are split by section headers, with each section becoming one chunk. The specific corpus used in this study (IfcOpenShell) is described in Section~\ref{subsec:implementation}.

At index time, each chunk undergoes LLM-based review that serves two purposes: filtering non-useful chunks (e.g., utilities, deprecated functions) and generating three to five \textit{reverse questions} per useful chunk, i.e., questions a user might ask that the chunk would answer. Following the document expansion by query prediction approach~\cite{nogueiraDocumentExpansionQuery2019}, this enables a third retrieval channel (Section~\ref{subsubsec:retrieval-pipeline}) that matches user queries against anticipated questions rather than raw documentation content, bridging the vocabulary gap between how users phrase questions and how documentation describes functionality. Implementation parameters (embedding model, corpus size) are specified in Section~\ref{subsec:implementation}.

\subsubsection{Retrieval Pipeline}
\label{subsubsec:retrieval-pipeline}

At query time, three parallel retrieval channels produce candidate sets: (1) dense chunk search via cosine similarity over chunk embeddings, (2) BM25 (Best Matching 25)~\cite{robertsonProbabilisticRelevanceFramework2009a} lexical search over a sparse index, and (3) dense reverse-question search that matches the query embedding against the generated question collection and maps results back to parent chunks. The hybrid design is motivated by Chen et al.~\cite{chenWhenLLMsMeet2025}, who found that BM25 outperforms dense retrievers for code-related tasks, while dense search captures semantic similarity that lexical matching misses.

The three ranked lists are combined via Reciprocal Rank Fusion (RRF)~\cite{cormackReciprocalRankFusion2009}, and the top-$n_r$ fusion candidates are reranked by a cross-encoder, a transformer model that jointly encodes the query and each candidate chunk to produce a fine-grained relevance score, yielding more accurate rankings than the embedding-based similarity used in the initial retrieval channels. The final top-$k$ chunks are formatted as numbered documentation blocks and injected into the agent's execution context. The complete retrieval algorithm is available in the project repository.

\subsection{Domain-Specific Tools}
\label{subsec:tools}

While the agent can write arbitrary code, domain-specific tools offer higher-level abstractions that encapsulate recurring extraction patterns and embed domain knowledge. By calling a tool, the agent avoids re-implementing common operations from scratch, potentially reducing code complexity and error rates. Tools represent the second augmentation axis in our experimental design (Section~\ref{sec:experimental-setup}), where we compare manual and automatically generated variants against a no-tool baseline.

\subsubsection{Manual Tool Development}
\label{subsubsec:manual-tools}

Manual tools are hand-crafted functions that encode domain expertise for common extraction tasks (e.g., retrieving element properties, computing spatial relationships). They serve as a human-engineered baseline against which automatically generated tools are compared. The specific tools used in this study are described in Section~\ref{subsec:implementation}.

\subsubsection{Automated Tool Generation}
\label{subsubsec:auto-tool-gen}

As an alternative to manual development, we introduce a training pipeline that automatically generates tools through a multi-agent system (MAS). This architecture is motivated by evidence that separating code generation from testing and validation across specialized agents reduces cascading errors in code synthesis tasks (Section~\ref{subsubsec:tool-generation}). Following this principle, the pipeline comprises seven specialized agents coordinating through deterministic control flow (Figure~\ref{fig:training-overview}). Each agent has a distinct role:

\begin{enumerate}
\item \textbf{Answer Generator}: Serves a dual role as the extraction system (identical to inference-time deployment) and as tool validator during training.
\item \textbf{Answer Verifier}: Classifies system answers against ground truth as correct, wrong, or abstained, following the LLM-as-judge paradigm~\cite{guSurveyLLMasaJudge2024}.
\item \textbf{Tool Identifier}: Analyzes execution histories from correct answers to identify reusable extraction patterns and recommends tool creation or enhancement.
\item \textbf{Error Analyst}: Diagnoses incorrect answers to determine whether a faulty tool caused the failure.
\item \textbf{Tool Creator}: Synthesizes tool implementations from identified patterns through iterative refinement.
\item \textbf{Tool Debugger}: Repairs faulty tool implementations using error descriptions and execution context.
\item \textbf{Tool Assessment}: Provides independent quality verification after tool creation or debugging.
\end{enumerate}

The state machine governing agent transitions is available in the project repository.

\subsubsection{Training Workflow}
\label{subsubsec:training-workflow}

\begin{figure}[htbp]
\centering
\includegraphics[width=\columnwidth]{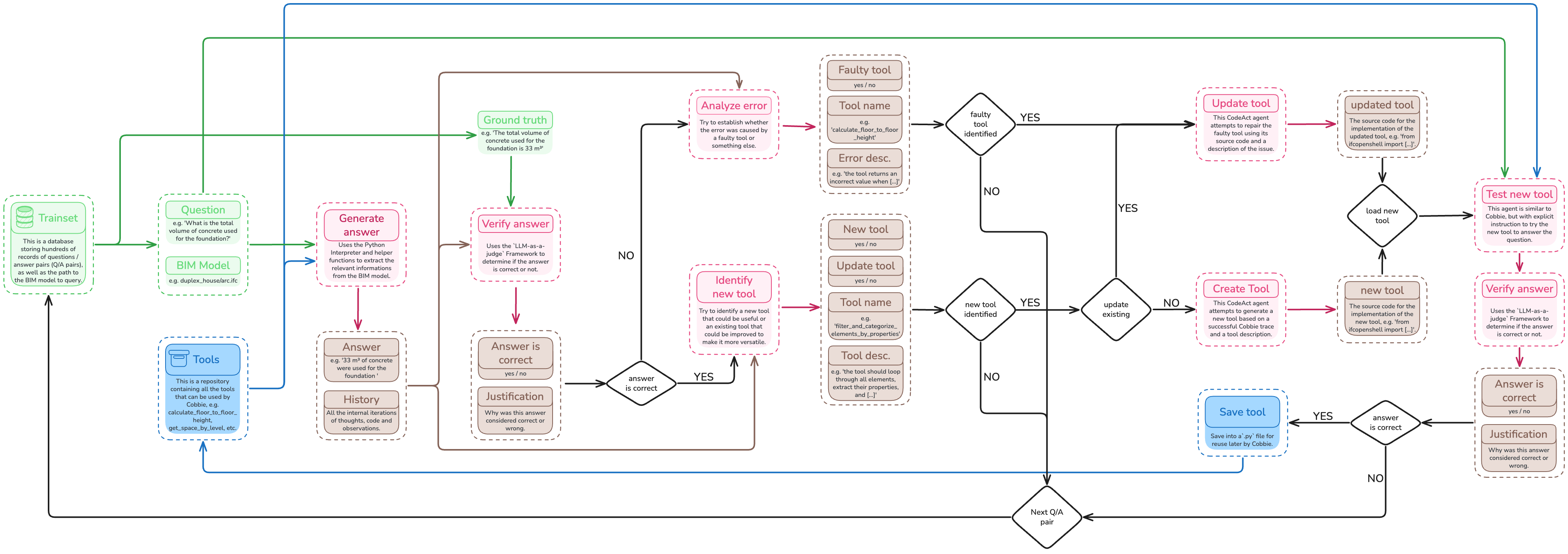}
\caption{Automated tool generation pipeline (optional augmentation). Path A (correct answers) identifies reusable patterns for new tools; Path B (wrong answers) diagnoses and repairs faulty tools. Both paths converge at testing and independent assessment before repository inclusion.}
\label{fig:training-overview}
\end{figure}

The training phase processes each tuple $(q, a^{gt}, m)$ from the train/dev partition through two convergent paths (Figure~\ref{fig:training-overview}). This partition was used for both the automated tool generation pipeline and all system development activities (prompt engineering, harness iteration). In \textbf{Path A (correct answer, tool creation)}, the Tool Identifier analyzes the execution history to identify reusable patterns, and the Tool Creator synthesizes a function implementation. In \textbf{Path B (wrong answer, tool debugging)}, the Error Analyst examines the execution history to identify faulty tools, and the Tool Debugger repairs the implementation. Both paths converge at a unified testing phase: the Answer Generator re-runs the question with the candidate tool, and the Tool Assessment Agent independently verifies whether the tool contributed to a correct answer. The complete state-based workflow is formalized in the project repository.

\subsubsection{Tool Testing, Validation, and Repository Management}
\label{subsubsec:tool-repo}

Tools must demonstrably improve extraction accuracy to be persisted. A bounded repository ($N_{max} = 16$) enforces evolutionary pressure through usage-based pruning. When the repository exceeds capacity, the tool with the highest deletion score is removed:
\begin{equation}
s_{del}(t) = \frac{1 - r_{call}(t)}{2} + \frac{1 - r_{succ}(t)}{2}
\end{equation}
where $r_{call}(t)$ is the fraction of questions where the tool was invoked (when available) and $r_{succ}(t)$ is the fraction of invocations that contributed to a correct answer. A grace period of $N_{grace}$ questions protects recently created tools from premature deletion before sufficient usage data accumulates. This mechanism ensures that only genuinely useful tools survive, analogous to evolutionary selection in self-improving systems~\cite{zhangDarwinGodelMachine2025, novikovAlphaEvolveCodingAgent2025}.

\section{Experimental Setup}
\label{sec:experimental-setup}

This section describes the experimental design used to evaluate the adaptive exploration paradigm and its interaction with augmentation strategies and model capability. All experiments access IFC files through IfcOpenShell~\cite{thomaskrijnenIfcOpenShell2025}, a widely used open-source Python library for parsing and manipulating IFC data. We first introduce the benchmark used for evaluation (Section~\ref{subsec:ifc-bench}), then describe the static baseline configuration (Section~\ref{subsec:static-baseline}), the factorial design, evaluation criteria, and statistical methods (Section~\ref{subsec:eval-framework}), and finally report implementation details (Section~\ref{subsec:implementation}).

\subsection{ifc-bench v2}
\label{subsec:ifc-bench}

As discussed in Section~\ref{subsubsec:benchmarks-eval}, existing BIM-QA datasets are too small and narrowly scoped to support systematic evaluation. We introduce ifc-bench v2, which addresses these limitations through three design goals: (1)~scale sufficient for statistically powered ablation studies (Section~\ref{subsubsec:benchmark-stats}), (2)~a formal question categorization taxonomy that enables per-category analysis, and (3)~open-ended questions requiring judgment and estimation, moving beyond the deterministically verifiable answers used in all prior BIM-QA datasets. Rather than optimizing for high accuracy on a narrow set of hand-picked queries, ifc-bench v2 prioritizes diversity and evaluation rigor, with substantial headroom for future methods to improve.

\subsubsection{Task Construction and Categorization}
\label{subsubsec:task-construction}

Each task consists of a natural-language question, a reference IFC model, and a ground-truth answer. Questions are categorized using a four-category taxonomy adapted from Solihin et al.~\cite{solihinClassificationRulesAutomated2015} and formalized in our evaluation framework~\cite{hellinsylvainSystematicEvaluationFramework2026}:
\begin{itemize}
\item \textbf{Category~1 -- Direct Information Retrieval:} Answers obtainable by looking up explicit properties or attributes (e.g., ``What is the fire rating of door D-201?'').
\item \textbf{Category~2 -- Computational Aggregation:} Answers requiring aggregation or counting across multiple elements (e.g., ``How many walls are on the second floor?'').
\item \textbf{Category~3 -- Geometric/Spatial Computation:} Answers requiring geometric calculations or spatial reasoning (e.g., ``What is the total window area on the south facade?'').
\item \textbf{Category~4 -- Incomplete Information Scenarios:} Questions where the required information is absent, ambiguous, or must be estimated from available data.
\end{itemize}

\subsubsection{Model Diversity and Selection}
\label{subsubsec:model-diversity}

The benchmark corpus spans 37 IFC models from 21 projects. Models were sourced from two origins: publicly available IFC repositories (e.g., buildingSMART sample files, KIT FZK House, open BIM datasets) and models created by students at the Technical University of Munich as part of BIM coursework projects. None of the models were authored or modified for this study. The corpus covers multiple authoring tools (Revit 2011--2025, ArchiCAD 11--25, Synchro, DDS-CAD), IFC schema versions (IFC2X3, IFC4, IFC4X3), building types (office, residential, healthcare, mixed-use), and languages. Table~\ref{tab:model_metadata} summarizes the corpus characteristics, including entity counts and validation issue counts per project.

Models exhibit varying levels of representational quality, as is typical for real-world BIM data. Validation issue counts, obtained by running each model through Solibri Model Checker\footnote{Solibri, \url{https://www.solibri.com/}} prechecks (model structure, component validity, clearance, deficiency detection, and space checks), range from 0 to over 4,000 per project. Common deficiencies include missing attributes, absent spatial elements (e.g., no \texttt{IfcSpace} definitions), inconsistent property sets, and misclassified components. This diversity is intentional; real-world BIM models rarely pass all validation checks, and a benchmark restricted to well-formed models would not test robustness to BIM heterogeneity.

\begin{table}[htbp]
\centering
\footnotesize
\caption{IFC model corpus: 21 projects spanning diverse authoring tools, schema versions, and building types. Element count and file size are summed across all model files per project. Projects marked with $\dagger$ were used for manual tool development.}
\label{tab:model_metadata}
\resizebox{\textwidth}{!}{%
\begin{tabular}{l l l l r r r r}
\hline
\textbf{Project} & \textbf{Authoring tool(s)} & \textbf{Schema} & \textbf{Type} & \textbf{Mod.} & \textbf{Elem.} & \textbf{MB} & \textbf{Iss.} \\
\hline
4351                     & Revit 2011           & 2X3     & other*      & 1 & 441      & 7.3    & 0 \\
ac20                     & ARCHICAD 20          & 4       & office      & 1 & 1,190    & 10.9   & 8 \\
city\_house\_munich      & Revit 2025           & 4X3     & living      & 1 & 293      & 33.5   & 2 \\
dental\_clinic$\dagger$  & Revit 2011, Solibri  & 2X3     & healthcare  & 3 & 20,940   & 158.4  & 270 \\
digital\_hub             & Revit 2019           & 4       & office      & 4 & 13,677   & 67.8   & 0 \\
duplex$\dagger$          & Revit 2011           & 2X3     & living      & 2 & 1,268    & 12.8   & 14 \\
ettenheim\_gis           & ADT IFC-Utility      & 2X3     & mix         & 2 & 3,777    & 61.6   & 0 \\
fantasy\_hotel\_1        & Revit 2024           & 4       & hospitality*& 1 & 1,082    & 8.5    & 25 \\
fantasy\_hotel\_2        & Revit 2024           & 4       & hospitality*& 1 & 1,571    & 10.0   & 118 \\
fantasy\_office\_1       & Revit 2025           & 4       & office*     & 1 & 1,846    & 13.4   & 62 \\
fantasy\_office\_2       & Revit 2025           & 4       & office*     & 1 & 1,644    & 19.5   & 0 \\
fantasy\_office\_3       & Revit 2024           & 4       & office*     & 1 & 1,460    & 9.7    & 48 \\
fantasy\_residential\_1  & Revit 2025           & 4       & living*     & 1 & 1,055    & 19.7   & 12 \\
fzk\_house               & ARCHICAD 20          & 4       & living      & 1 & 127      & 2.6    & 5 \\
hitos                    & ArchiCAD 11          & 2X3     & office      & 1 & 12,148   & 64.1   & 4,262 \\
molio                    & ARCHICAD 25          & 2X3     & office      & 1 & 3,210    & 74.0   & 2,862 \\
samuel\_macalister       & Revit 2014, DDS-CAD  & 4       & living      & 2 & 7,014    & 55.0   & 202 \\
schependomlaan           & Synchro 5.0          & 2X3     & living      & 1 & 3,508    & 65.1   & 74 \\
sixty5                   & ARCHICAD 21, Revit 2018 & 2X3  & living      & 7 & 188,312  & 839.9  & 321 \\
smiley\_west             & ARCHICAD 20          & 4       & living      & 1 & 1,272    & 6.1    & 122 \\
wbdg\_office             & Revit 2011           & 2X3     & office      & 3 & 7,498    & 57.1   & 90 \\
\hline
\textbf{Total}           &                      &         &             & \textbf{37} & \textbf{273,333} & \textbf{1,596.8} & \textbf{8,497} \\
\hline
\end{tabular}}%
\begin{flushleft}
\footnotesize
\textit{Note:} Schema abbreviations: 2X3 = IFC2X3, 4 = IFC4, 4X3 = IFC4X3. Mod.\ = number of model files. Elem.\ = total IFC entity count. Iss.\ = validation issues from Solibri Model Checker prechecks (model structure, component validity, clearance, deficiency detection, and space checks). Types marked with * are inferred from project naming and context rather than explicit model metadata.
\end{flushleft}
\end{table}

\begin{figure}[htbp]
\centering
\includegraphics[width=\textwidth]{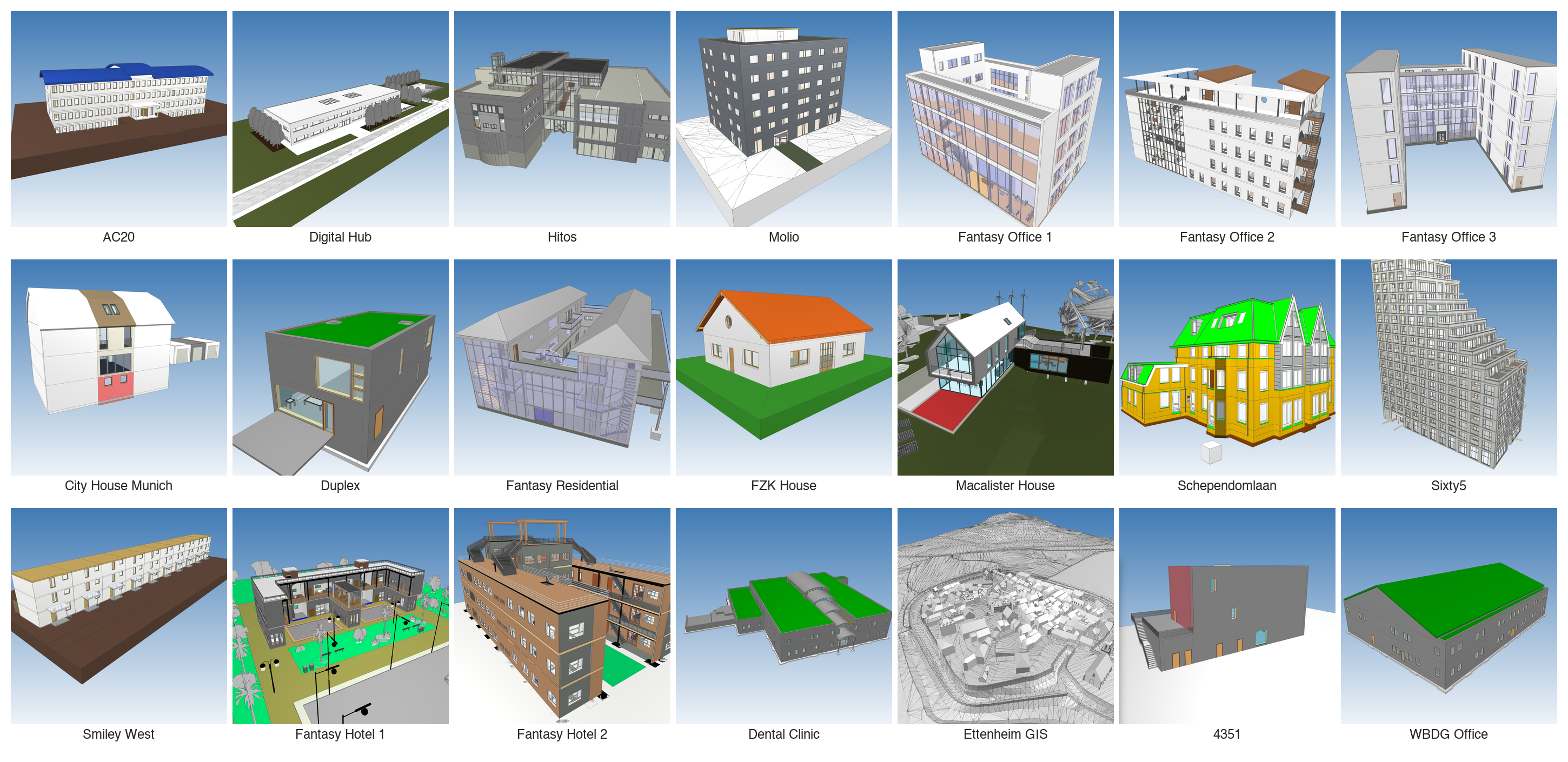}
\caption{Representative 3D views of the 21 projects in the ifc-bench corpus, illustrating diversity in building type, scale, and geometric complexity.}
\label{fig:bim_model_grid}
\end{figure}

\subsubsection{Benchmark Statistics and Comparison to Prior Datasets}
\label{subsubsec:benchmark-stats}

ifc-bench v2 comprises 1,027 question-answer pairs across 21 projects and 37 IFC models, split into a train/dev partition (513) and a test partition (514) via 50\% stratified split (seed~=~42). The train/dev partition was used for both the automated tool generation pipeline and system development (prompt engineering, harness iteration); all reported results come exclusively from the held-out test partition. Compared to ifc-bench v1~\cite{hellinNaturalLanguageInformation2025}, v2 represents an order-of-magnitude expansion across all dimensions (questions, projects, and models). This scale gap is even larger relative to other BIM-QA datasets: Koh et al.~\cite{kohCosteffectiveMinimalinterventionBIM2026} and Guo et al.~\cite{guoAdvancingBIMInformation2025} use 80 queries on 1--2 models; Liu et al.~\cite{liuIntegratedMethodBIM2025} use 40 queries on a single model; Wei et al.~\cite{weiTexttostructureInterpretationUser2025} use 11 questions. The benchmark and evaluation tools are released under a permissive open-source license.

\subsection{Static Baseline Configuration}
\label{subsec:static-baseline}

The static baseline isolates iterative interaction as the single variable under test. In a zero-shot configuration, the same LLM receives the question, model path, and any available tools or documentation, then generates a single code block. This code is executed once; the LLM receives the combined stdout and stderr and produces a final answer or abstains. Unlike the adaptive configuration, the LLM cannot issue follow-up code. This configuration mirrors Austern et al.~\cite{austernComparingDifferentBuilding2025}, who generate code and execute it in a single pass.

For the documentation-augmented static variant, a three-step pipeline pre-fetches relevant documentation before code generation. First, a query planner generates up to five documentation query strings based on the question. Second, the same hybrid retrieval pipeline used in the adaptive configuration (Section~\ref{subsec:doc-retrieval}) fetches and deduplicates chunks, capped at 10 chunks. Third, the code generator receives the pre-fetched documentation as a prompt section and generates a single code block as above.

\subsection{Evaluation Framework}
\label{subsec:eval-framework}

\subsubsection{Ablation Matrix Design}
\label{subsubsec:ablation-design}

The experiment follows a 3$\times$4 factorial design crossing three LLM configurations (rows) with four augmentation strategies (columns), yielding 12 cells:
\begin{itemize}
\item \textbf{Rows} represent model configurations that jointly determine the exploration paradigm and model capability: \emph{adaptive-4.7} (iterative exploration, full-size model), \emph{adaptive-4.5~Air} (iterative exploration, reduced capability), and \emph{static-4.7} (single-pass code generation, full-size model).
\item \textbf{Columns} represent augmentation strategies applied independently: \emph{none} (vanilla), \emph{documentation} (IfcOpenShell API retrieval, Section~\ref{subsec:doc-retrieval}), \emph{manual tools} (hand-crafted extraction functions), and \emph{auto tools} (machine-generated tools, Section~\ref{subsubsec:auto-tool-gen}).
\end{itemize}
\noindent Row comparisons (adaptive vs.\ static, holding augmentation constant) address RQ1. The full matrix addresses RQ2 by systematically varying augmentation and model capability.

\subsubsection{Evaluation Criteria}
\label{subsubsec:eval-criteria}

Each answer is evaluated against five binary criteria from our evaluation framework~\cite{hellinsylvainSystematicEvaluationFramework2026}: (0)~\emph{abstention}: did the system provide an answer; (1)~\emph{faithfulness}: are claims grounded in acceptable sources; (2)~\emph{completeness}: all relevant facts included; (3)~\emph{transparency}: sources and methods disclosed; and (4)~\emph{relevance}: directly addresses the question. Each criterion is evaluated as a binary judgment (true / false) per answer. Faithfulness thresholds are category-dependent: Category~1 requires grounding in BIM data only, while Category~4 permits stated assumptions alongside data evidence.

Abstention occurs when the agent explicitly decides it cannot answer, or when its iteration budget is exhausted (i.e., it returns the default ``Information not found in BIM model'' message from Algorithm~\ref{alg:inference}). This is distinct from system errors (crashes, timeouts), which are excluded from evaluation. The abstention rate is computed as:
\begin{equation}
r_{\text{abs}} = \frac{|\{i : \text{abstained}_i\}|}{N}
\end{equation}

The primary metric reported throughout this study is \emph{aggregate accuracy}: a non-abstained answer is rated \emph{correct} only when all four substantive criteria (1--4) are simultaneously satisfied:
\begin{equation}
\text{accuracy} = \frac{|\{i : c_{1,i} \wedge c_{2,i} \wedge c_{3,i} \wedge c_{4,i}\}|}{N}
\end{equation}
This metric is deliberately conservative: a single criterion failure marks the answer incorrect. We additionally report per-criterion pass rates (faithfulness and completeness in Table~\ref{tab:quality_matrix}) and accuracy among attempted answers (correct/attempted) as complementary perspectives. At scale, evaluation is automated using an LLM-as-judge protocol validated in our companion study~\cite{hellinsylvainENABLINGCROSSSTUDYCOMPARISON2026}, where LLM judges achieved higher inter-rater reliability ($\alpha = 0.70$--$1.00$, Krippendorff's alpha) than human experts ($\alpha = 0.32$--$0.57$) on the same tasks.

\subsubsection{Statistical Analysis}
\label{subsubsec:statistical-analysis}

Accuracy is reported with bootstrap 95\% confidence intervals (10,000 resamples). Pairwise comparisons use McNemar's test on paired data, with exact binomial computation when discordant pairs fall below 25. Significance levels: $^{*}p < 0.05$, $^{**}p < 0.01$, $^{***}p < 0.001$. Per-category results with $n < 50$ are described as indicative.

\subsection{Implementation Details}
\label{subsec:implementation}

All experiments use GLM backbone models from Zhipu AI\footnote{Zhipu AI (Beijing Zhipu Huazhang Technology Co., Ltd.), \url{https://www.zhipuai.cn/}}. The full-capability configuration uses GLM~4.7 (355B total parameters, 32B active via sparse mixture-of-experts with 160 routed experts and 8 active per token), while the reduced-capability configuration uses GLM~4.5~Air (106B total, 12B active via MoE with 128 experts and 9 active per token), providing approximately 2.7$\times$ fewer active parameters. Both models access IFC files through IfcOpenShell~\cite{thomaskrijnenIfcOpenShell2025}, with code execution occurring in persistent, sandboxed Python environments with network access disabled.

The system prompt directs the agent to use IfcOpenShell directly as well as pre-loaded helper functions (when applicable), preferring helper functions over direct API calls when available. Claims must be sourced through a four-tier hierarchy: direct retrieval from BIM properties, simple computation (counting, summing), complex computation (geometric/spatial calculations), and stated assumptions when information is unavailable. The agent's responses are evaluated against five quality criteria (faithfulness, completeness, transparency, relevance, and appropriate abstention) that are embedded directly in the prompt.

The evaluation covers all 12 configurations of the ablation matrix, each evaluated on the held-out test partition of ifc-bench (514 question-answer pairs). The category distribution in the test set is: Category~1~=~72, Category~2~=~283, Category~3~=~57, and Category~4~=~102.
\section{Results}
\label{sec:results}

This section presents results organized by research question. Section~\ref{subsec:rq1} compares adaptive and static exploration paradigms, while Section~\ref{subsec:rq2} examines how augmentation strategies and model capability interact across the full ablation matrix.

\subsection{RQ1: Adaptive vs.\ Static Exploration}
\label{subsec:rq1}

Adaptive exploration outperforms single-pass code generation across both model capability levels and all valid augmentation configurations. The paradigm gap dominates all other factors in the ablation.

\subsubsection{Overall Accuracy Comparison}
\label{subsubsec:rq1-overall}

Table~\ref{tab:accuracy_matrix} reports the full 3$\times$4 accuracy matrix. Adaptive-4.7 outperforms Static-4.7 by +36.8--38.5 pp across all completed augmentation strategies (all $p < 0.001$, McNemar's test). This gap dwarfs the effect of any augmentation strategy: the largest significant augmentation effect is +4.9pp (documentation on GLM~4.5~Air), roughly one-eighth the paradigm gap.

The paradigm advantage extends to the weaker model. Even with approximately 2.7$\times$ fewer active parameters and no augmentation, Adaptive-4.5~Air outperforms Static-4.7 ($p < 0.01$), demonstrating that iterative exploration compensates for reduced model capability.

Abstention rates further distinguish the paradigms. Static-4.7 abstains on roughly half the benchmark even when tools are available, while Adaptive-4.7 abstains on fewer than 7\% of questions. Among attempted answers, Adaptive-4.7 also achieves a higher success rate (60.1\% vs.\ 42.6\%). The adaptive paradigm thus dominates on both dimensions: it attempts far more questions and succeeds on a higher fraction of those it attempts.

\begin{table}[htbp]
\centering
\small
\caption{Accuracy across the 3$\times$4 ablation matrix with bootstrap 95\% confidence intervals. Rows represent augmentation strategies; columns represent model configurations. Accuracy is the proportion of answers where all four evaluation criteria are satisfied. Bold indicates the highest accuracy per row.}
\label{tab:accuracy_matrix}
\begin{tabular}{l c c c}
\hline
\textbf{Augmentation} & \textbf{Adaptive (4.7)} & \textbf{Adaptive (4.5 Air)} & \textbf{Static (4.7)} \\
\hline
None   & \textbf{56.0\%} [51.8, 60.3] & 25.7\% [22.0, 29.4] & 19.1\% [15.8, 22.5] \\
Doc    & \textbf{56.6\%} [52.3, 60.9] & 30.6\% [26.7, 34.7] & 20.6\%$^\dagger$ \\
Manual & \textbf{55.4\%} [51.2, 59.7] & 15.4\% [12.4, 18.5] & 18.6\% [14.0, 23.3] \\
Auto   & \textbf{55.4\%} [51.2, 59.7] & 23.8\% [20.2, 27.4] & 16.9\% [13.5, 20.2] \\
\hline
Abst.  & 5.8--6.8\% & 16.3--50.6\% & 43.0--55.1\% \\
\hline
\end{tabular}
\begin{flushleft}
\footnotesize
\textit{Note:} 95\% confidence intervals computed via 10,000 bootstrap resamples. Sample sizes: $n=514$ per cell. Abstention rate ranges reflect the minimum and maximum across model configurations per row.
\end{flushleft}
\end{table}

\subsubsection{Per-Category Breakdown}
\label{subsubsec:rq1-categories}

The paradigm advantage varies systematically by category (Table~\ref{tab:per_category}). Category~3 (Geometric/Spatial Computation) shows the largest gap: the static configuration abstains on over 75\% of these questions, effectively refusing computation tasks when no pre-computed property is available. Adaptive exploration enables the agent to iteratively derive geometric quantities across multiple execution rounds.

A notable finding is that Category~3 accuracy (66.7\%) exceeds Category~1 (55.6\%) for the adaptive configuration, inverting the expected ordering where direct property retrieval should be easier than computation. This inversion has three root causes: materials-related questions (26\% of Category~1) achieve only 42\% accuracy due to deeply nested IFC material structures that the agent systematically under-explores; Category~1 frequently requires exhaustive element lists where any omission triggers failure, while Category~3 answers are predominantly single computed values; and the adaptive paradigm disproportionately helps computation (+61.4pp over static) versus retrieval (+31.6pp). Under the static paradigm, the expected ordering holds (Cat~1: 23.9\% $>$ Cat~3: 5.3\%). We discuss the implications of this finding in Section~\ref{subsec:why-dynamic}.

Categories~1 and 2 (Direct Retrieval, Aggregation) show consistent adaptive advantages for both models. Category~4 (Incomplete Information Scenarios) is the sole exception: Adaptive-4.5~Air underperforms Static-4.7 on these tasks. This likely reflects the stronger model's broader world knowledge, enabling it to make informed assumptions, a response type acceptable under Category~4 evaluation criteria when assumptions are explicitly stated. Adaptive-4.7 still outperforms Static-4.7 on Category~4, confirming this reflects a model capability floor rather than a paradigm limitation.

\begin{table}[htbp]
\centering
\small
\caption{Per-category accuracy (correct rate, \%) across the 3$\times$4 ablation matrix. Bold indicates the highest accuracy per category. Category 3 has $n < 60$ per cell; results should be interpreted with caution.}
\label{tab:per_category}
\resizebox{\textwidth}{!}{%
\begin{tabular}{l l r r r r}
\hline
\textbf{Category} & \textbf{Configuration} & \textbf{None} & \textbf{Doc} & \textbf{Manual} & \textbf{Auto} \\
\hline
1 -- Direct Property  & Adaptive (4.7)     & 55.6 & \textbf{63.9} & 59.7 & 58.3 \\
($n=72$)              & Adaptive (4.5 Air) & 31.9 & 27.4 & 23.3 & 21.6 \\
                      & Static (4.7)      & 23.9 & $\dagger$  & 20.0 & 12.5 \\
\hline
2 -- Aggregation      & Adaptive (4.7)     & \textbf{53.4} & 51.6 & 49.8 & 50.5 \\
($n=283$)             & Adaptive (4.5 Air) & 20.4 & 27.1 & 13.0 & 21.6 \\
                      & Static (4.7)      & 15.2 & $\dagger$  & 14.2 & 13.6 \\
\hline
3 -- Computation      & Adaptive (4.7)     & 66.7 & \textbf{68.4} & 66.7 & 66.7 \\
($n=57$)              & Adaptive (4.5 Air) & 36.8 & 42.1 & 24.6 & 27.6 \\
                      & Static (4.7)      &  5.3 & $\dagger$  &  3.8 &  3.5 \\
\hline
4 -- Estimation/Unav. & Adaptive (4.7)     & 57.8 & 58.8 & \textbf{61.8} & 60.8 \\
($n=102$)             & Adaptive (4.5 Air) & 29.4 & 36.3 & 11.7 & 29.0 \\
                      & Static (4.7)      & 34.3 & $\dagger$  & 35.7 & 36.3 \\
\hline
\end{tabular}}%
\begin{flushleft}
\footnotesize
\textit{Note:} Accuracy is the correct rate (correct answers / total $n$), consistent with Table~\ref{tab:accuracy_matrix}. The $n$ shown is per cell within each category.
\end{flushleft}
\end{table}

\subsubsection{Qualitative Analysis}
\label{subsubsec:rq1-qualitative}

Three examples illustrate the core mechanisms of adaptive exploration. In a Category~3 (Computation) task, the agent was asked for the total volume of concrete used in walls (project 4351), a quantity absent from the model's explicit properties. Over 13 iterations, the agent analyzed \texttt{IfcMaterialLayerSetUsage} to identify concrete layers, computed geometric volumes, and applied layer-thickness ratios to isolate concrete fractions, recovering from multiple unsuccessful IfcOpenShell API calls before converging. The static configuration abstained. In a Category~2 (Aggregation) task, the agent encountered non-standard level naming (``OK OG2'' for Level~2) that no static lookup would find. In the adaptive configuration, the agent iteratively explored spatial containment hierarchies, discovered the naming convention, and located the relevant elements after several failed attempts at standard property paths.

A Category~4 (Incomplete Information) task asked ``What type of roof structure is used?'' for project 4351, where no explicit roof-type property is stored. Over 14 iterations, the agent assembled indirect evidence: it identified a parapet wall at 46\,ft elevation, located a roof soffit element, and confirmed the absence of any pitched or trussed structural members, ultimately inferring a flat reinforced concrete slab roof. The static configuration abstained entirely, as no single property lookup could answer the question.

These examples represent the three core mechanisms: runtime data structure discovery, error recovery via execution feedback, and multi-step synthesis from circumstantial evidence. Representative examples for all four categories are available in the project repository.

\subsection{RQ2: Augmentation Strategy and Model Capability Interaction}
\label{subsec:rq2}

The full 3$\times$4 matrix reveals that augmentation effects are strongly model-capability-dependent. The high-capability model is invariant to augmentation; the lower-capability model responds differently to each strategy.

\subsubsection{Main Results}
\label{subsubsec:rq2-main}

GLM~4.7 accuracy spans 55.4--56.6\% across all four augmentation strategies: a 1.2pp range with no significant pairwise differences (all $p > 0.8$). GLM~4.5~Air spans 15.4--30.6\%: a 15.2pp range with several significant pairwise differences. The same augmentation strategies that leave the stronger model unchanged produce large positive and negative effects on the weaker model.

We focus the quality analysis on faithfulness and completeness because the remaining criteria (transparency and relevance) approach ceiling for all configurations: GLM-4.7 transparency ranges 94.2--97.7\% and relevance 96.8--98.5\%, leaving insufficient variance for meaningful comparison. Quality metrics for the discriminating criteria (Table~\ref{tab:quality_matrix}) follow the same pattern as aggregate accuracy. Adaptive-4.7 faithfulness (65.5--68.1\%) and completeness (81.5--82.6\%) are stable across augmentations. Static-4.7 achieves lower quality on both dimensions (faithfulness 51.7--59.5\%, completeness 52.3--54.3\%); despite high abstention selecting only easier questions, Static-4.7 still underperforms. Adaptive-4.5~Air shows moderate quality (faithfulness 50.0--52.5\%, completeness 62.2--69.9\%).

\begin{table}[htbp]
\centering
\small
\caption{Answer quality across the 3$\times$4 ablation matrix: faithfulness and completeness rates for non-abstained answers. Bold indicates the highest rate per column within each metric. GLM-4.7 transparency (94.2--97.7\%) and relevance (96.8--98.5\%) remain near-ceiling; GLM-4.5 Air shows lower transparency (64.2--69.7\%) and relevance (92.6--95.0\%).}
\label{tab:quality_matrix}
\begin{tabular}{l c c c c}
\hline
 & \textbf{None} & \textbf{Doc} & \textbf{Manual} & \textbf{Auto} \\
\hline
\multicolumn{5}{l}{\textit{Faithfulness}} \\
\quad Adaptive (4.7)     & \textbf{68.1\%} & \textbf{66.8\%} & \textbf{65.5\%} & \textbf{66.8\%} \\
\quad Adaptive (4.5 Air) & 50.0\%          & 51.0\%          & 52.5\%          & 52.3\% \\
\quad Static (4.7)      & 59.5\%          & $\dagger$       & 52.0\%          & 51.7\% \\
\hline
\multicolumn{5}{l}{\textit{Completeness}} \\
\quad Adaptive (4.7)     & \textbf{81.5\%} & \textbf{82.0\%} & \textbf{81.8\%} & \textbf{82.6\%} \\
\quad Adaptive (4.5 Air) & 62.2\%          & 67.1\%          & 69.3\%          & 69.9\% \\
\quad Static (4.7)      & 52.9\%          & $\dagger$       & 52.3\%          & 54.3\% \\
\hline
\end{tabular}
\begin{flushleft}
\footnotesize
\textit{Note:} Rates computed over non-abstained answers only. Static configurations show lower effective sample sizes due to high abstention rates (43.0--55.1\%).
\end{flushleft}
\end{table}

\subsubsection{Effect of Documentation Retrieval}
\label{subsubsec:rq2-docs}

For GLM~4.7, documentation retrieval has no measurable effect. Accuracy is 56.6\% with documentation versus 56.0\% without (+0.6pp, $p = 0.86$, ns). The model's iterative execution loop provides sufficient information discovery, rendering pre-loaded API documentation redundant.

For GLM~4.5~Air, documentation retrieval produces a statistically significant improvement. Accuracy increases from 25.7\% to 30.6\% (+4.9pp; bootstrap 95\% CI [+0.6, +9.6pp], $p = 0.036$). Abstention falls from 20.2\% to 16.3\%, suggesting that the information scaffold reduces exhaustion of the iteration budget. The weaker model benefits from pre-loaded context it cannot independently generate through iteration.

This divergence is notable given strong prior evidence that documentation benefits code generation for less common Python libraries~\cite{chenWhenLLMsMeet2025}, conditions directly applicable to IfcOpenShell. For the weaker model, the documentation system functions as the literature predicts. For the stronger model, the iterative execution loop substitutes for explicit documentation, yielding a null result.

\subsubsection{Effect of Domain-Specific Tools}
\label{subsubsec:rq2-tools}

For the adaptive method with GLM~4.7, neither manual nor auto-generated tools produce significant changes. Manual tools yield $-0.6$pp ($p = 0.86$, ns); auto tools yield $-0.6$pp ($p = 0.86$, ns). The model generates effective exploration code without domain-specific abstractions.

For GLM~4.5~Air, manual tools are harmful. Accuracy drops from 25.7\% to 15.4\% ($-10.2$pp, $p < 0.001$), and abstention reaches 50.6\%. Execution trace inspection reveals a consistent failure mode: the model repeatedly attempts tool invocations but fails to make progress, exhausting the iteration budget before producing an answer. Tools designed to simplify extraction instead channel the weaker model into unproductive exploration loops.

Auto-generated tools outperform manual tools for GLM~4.5~Air by +8.3pp ($p < 0.001$), but remain slightly below the no-augmentation baseline ($-1.9$pp, $p = 0.45$, ns). Auto tools match or exceed manual accuracy while eliminating the development effort required for manual tool creation; a clear cost-effectiveness advantage, though neither strategy improves over the no-tool baseline for either model.

\subsubsection{Model-Capability Dependence}
\label{subsubsec:rq2-capability}

The augmentation results form a coherent interaction pattern. For GLM~4.7, all four augmentation strategies are effectively neutral (range: 55.4--56.6\%, all $p > 0.8$). For GLM~4.5~Air, documentation helps (+4.9pp, $p < 0.05$), manual tools hurt ($-10.2$pp, $p < 0.001$), and auto tools are marginally negative but non-significant ($-1.9$pp, $p = 0.45$).

Documentation provides an information scaffold that the weaker model cannot generate internally but the stronger model derives through iteration. Manual tools impose action constraints the weaker model cannot navigate, exhausting its iteration budget in unproductive loops. This pattern has a direct practical implication: augmentation investment is warranted only for weaker models (documentation only), not for high-capability models. The paradigm gap (Section~\ref{subsubsec:rq1-overall}) consistently exceeds augmentation effects by an order of magnitude; augmentation delivers at most +4.9pp under specific capability conditions.

\section{Discussion}
\label{sec:discussion}

The results establish that the adaptive exploration paradigm significantly outperforms static approaches in BIM information extraction, and identify it as the most influential factor in extraction accuracy. The paradigm advantage (+36.8--38.5pp, all $p < 0.001$) exceeds any augmentation effect by an order of magnitude, and even a weaker model with adaptive exploration outperforms the stronger model under static generation. This section first interprets the mechanisms underlying this advantage (Section~\ref{subsec:why-dynamic}), then examines augmentation effects and their practical implications (Section~\ref{subsec:augmentation-discussion}), and finally discusses limitations (Section~\ref{subsec:limitations}).

\subsection{Mechanisms Underlying the Paradigm Advantage}
\label{subsec:why-dynamic}

Two complementary mechanisms explain why adaptive exploration consistently outperforms static approaches. First, \textit{runtime data structure discovery}: rather than assuming a known schema, the agent inspects the BIM model to determine how information is organized. This is the dominant mechanism in Category~3, where static approaches abstain 75.4\% of the time because no pre-computed geometric property exists; the agent computes volumes through iterative schema exploration. Second, \textit{execution feedback} enables fault tolerance: when an extraction attempt fails or returns unexpected data, the agent adapts its strategy and recovers over multiple rounds. The Reflexion~\cite{shinnReflexionLanguageAgents2023} and Self-Refine~\cite{madaanSelfRefineIterativeRefinement2023} literatures formalize this mechanism.

Cross-model evidence validates the core claim. The paradigm advantage holds at both capability levels, and Category~3 shows large gains for both models. The Category~4 exception for the weaker model reflects a capability floor, not a paradigm limitation.

The Category~3 $>$ Category~1 accuracy inversion (Section~\ref{subsubsec:rq1-categories}) reveals that what constitutes ``difficulty'' for an LLM-based extraction agent differs fundamentally from human intuition. The ``Direct Property'' label masks a spectrum from trivial single-value lookups to exhaustive multi-element material decompositions. The adaptive paradigm is transformative for computation (+61.4pp over static) but only partially helps exhaustive retrieval (+31.6pp), because the agent systematically under-explores deeply nested structures such as \texttt{IfcMaterialLayerSetUsage}. This suggests that future improvements should target exhaustive traversal strategies for complex IFC structures.

Importantly, the paradigm advantage is unlikely to be specific to this implementation. The harness used here (Python with IfcOpenShell, CodeAct architecture) is one instantiation; alternative approaches such as MCP-based tool servers, SQL or Cypher queries, or CLI-based interaction would differ in absolute accuracy but preserve the core mechanism: iterative execution with runtime feedback will outperform single-pass code generation.

\subsection{Augmentation Effects and Practical Implications}
\label{subsec:augmentation-discussion}

Augmentation effects are modest and model-capability-dependent. Documentation retrieval benefits the weaker model (+4.9pp, $p < 0.05$) but yields no measurable improvement for the stronger model. One explanation is that augmentation may redirect exploration: when documentation is available, the agent may anchor on provided resources rather than exploring the model's actual data structures, reducing the diversity of exploration strategies. This null result, however, was observed for IfcOpenShell, a mature and well-documented library likely well-represented in pre-training data; less familiar or rapidly evolving libraries may still benefit from documentation retrieval even with capable models.

Domain-specific tools fail to improve accuracy for the stronger model and are actively harmful for the weaker one: manual tools reduce accuracy by 10.2pp ($p < 0.001$) as the model exhausts its iteration budget in unproductive tool invocation loops. Auto-generated tools match or outperform manual tools at zero development cost, provided a dataset exists to run the training pipeline, but neither strategy improves over the no-tool baseline. This pattern reflects a structural limitation rather than a failure of tool design: tools encode assumptions about data organization that inevitably break across the diverse conventions found in real-world models. Adaptive exploration succeeds precisely because it makes no such assumptions.

This observation also points to the potential value of standardized information requirements in BIM practice: specifications formalized through mechanisms such as Information Delivery Specifications (IDS) or Model View Definitions (mvdXML) constrain how data is modeled, which could reduce the representational heterogeneity that makes tool-based approaches brittle. Preliminary evidence from the benchmark is suggestive: the TUM coursework models, which were checked with Solibri and corrected as part of the curriculum, yield notably higher per-project accuracy than most other models of comparable size, suggesting that representational consistency may be a meaningful factor in extraction performance.

Although tools did not improve open-ended BIM queries, where heterogeneity is maximal, the automated tool generation pipeline itself is functional and produces syntactically correct, executable tools. In domains with more predictable structure, such as automated code compliance checking where expected elements and properties follow regulatory standards, the same pipeline could generate more effective tools. The pipeline's value thus depends on the structural consistency of the target domain, necessary to ensure deterministic outcomes from the generated tools, rather than the mechanism itself.

The model-capability interaction resonates with Sutton's bitter lesson~\cite{suttonBitterLesson2019}: methods that leverage general computation (iterative code execution) tend to outperform methods that encode human knowledge (tools, API documentation) as model capabilities increase. Two data points from one model family are suggestive, not conclusive, but practitioners should consider this trajectory when planning augmentation infrastructure.

\subsection{Limitations}
\label{subsec:limitations}

Several limitations bound the interpretation. \textit{Accuracy ceiling}: 55--57\% for the best configuration is insufficient for unsupervised deployment in safety-critical workflows. \textit{Format scope}: experiments are limited to IFC; cross-format generalization (Revit RVT, Archicad PLN, etc.) is untested. \textit{Model coverage}: both models tested are from the GLM family, and the model-dependent augmentation pattern is observed at two capability levels only; cross-family replication across diverse LLM providers and architectures, and extension to much weaker or substantially stronger models, are needed to confirm generalizability. \textit{Single-run evaluation}: although the large sample size (514 questions per configuration) and bootstrap confidence intervals mitigate sampling uncertainty, formal stability analysis across repeated runs was not performed. \textit{Benchmark representativeness}: ifc-bench v2 covers 37 IFC models from 21 projects and may not represent the full diversity of real-world BIM data. \textit{BIM model quality}: none of the benchmark models were validated against formalized information requirements (e.g., IDS or mvdXML specifications); without formalized compliance data, the relationship between model quality and extraction accuracy cannot be quantified in this study. \textit{Tool generalization}: domain-specific tools encode assumptions about data organization that limit their transferability across heterogeneous models. \textit{Library familiarity}: the documentation augmentation results are specific to IfcOpenShell, a stable, open-source library with substantial pre-training representation. The null effect for high-capability models may not generalize to less established or rapidly evolving APIs.

\section{Conclusion}
\label{sec:conclusion}

This work establishes adaptive exploration as the primary determinant of BIM information extraction accuracy. Across a factorial experiment comparing adaptive and static exploration paradigms at two model capability levels with four augmentation strategies, the paradigm gap (approximately $+$37--39pp within the high-capability model, $p < 0.001$) consistently exceeds augmentation effects ($\leq$4.9pp) by an order of magnitude. Even the weaker model with adaptive exploration outperforms the high-capability model with static code generation by $+$6.5pp, underscoring that paradigm choice matters more than raw model capability. For high-capability models, augmentation adds cost without measurable benefit. To support reproducibility and future research, ifc-bench v2 (1,027 question-answer pairs across 37 IFC models from 21 projects)\footnote{\url{https://huggingface.co/datasets/sylvainHellin/ifc-bench}} and the complete project repository\footnote{\url{https://github.com/sylvainHellin/cobbie}} are released under a CC-BY-4.0 license.

Key directions for future work include cross-family replication across diverse LLM providers and architectures, application of the tool generation pipeline to more structured domains such as code compliance checking, systematic analysis of how BIM model quality (e.g., compliance with formalized information requirements) affects extraction accuracy given the wide per-project spread (25--90\%, available in the project repository), and investigation of whether enforcing information delivery specifications narrows the gap between static and adaptive approaches.


\section*{CRediT Authorship Contribution Statement}
\textbf{S.~Hellin:} Conceptualization, Methodology, Software, Validation, Formal Analysis, Investigation, Data Curation, Writing -- Original Draft, Visualization.
\textbf{S.~Jang:} Conceptualization, Formal Analysis, Writing -- Review \& Editing.
\textbf{S.~Fuchs:} Writing -- Review \& Editing.
\textbf{S.~Nousias:} Supervision.
\textbf{A.~Borrmann:} Methodology, Writing -- Review \& Editing, Supervision, Funding Acquisition.

\section*{Declaration of Competing Interests}
The authors declare that they have no known competing financial interests or personal relationships that could have appeared to influence the work reported in this paper.

\section*{Data Availability}
The benchmark dataset (ifc-bench v2) is available at \url{https://huggingface.co/datasets/sylvainHellin/ifc-bench}. The project repository, including all source code and supplementary materials, is available at \url{https://github.com/sylvainHellin/cobbie}. Both are released under a CC-BY-4.0 license.

\section*{Declaration of Generative AI and AI-Assisted Technologies in the Manuscript Preparation Process}
During the preparation of this work, the authors used Claude Opus 4.6 (Anthropic) to improve readability and language quality (grammar, syntax, phrasing). After using this tool, the authors reviewed and edited the content as needed and take full responsibility for the content of the published article.

\section*{Acknowledgments}
This research was supported by the Georg Nemetschek Institute (GNI) and the Leonhard Obermeyer Center (LOC) at the Technical University of Munich.

\bibliographystyle{elsarticle-num}
\bibliography{bibliography/references}

\appendix

\section{Supplementary Materials}
\label{appendix:supplementary}

The project repository contains the following supplementary materials: the complete documentation retrieval algorithm, the training phase algorithm and state machine for automated tool generation, representative ifc-bench task examples, auto-generated tool implementations, and the full system prompt used for the answer generator.

\end{document}